# Further Exploration of the Dendritic Cell Algorithm: Antigen Multiplier and Time Windows

Feng Gu, Julie Greensmith and Uwe Aickelin

School of Computer Science, University of Nottingham, UK
{fxg,jqg,uxa}@cs.nott.ac.uk

Abstract. As an immune-inspired algorithm, the Dendritic Cell Algorithm (DCA), produces promising performance in the eld of anomaly detection. This paper presents the application of the DCA to a standard data set, the KDD 99 data set. The results of different implementation versions of the DCA, including antigen multiplier and moving time windows, are reported. The real-valued Negative Selection Algorithm (NSA) using constant-sized detectors and the C4.5 decision tree algorithm are used, to conduct a baseline comparison. The results suggest that the DCA is applicable to KDD 99 data set, and the antigen multiplier and moving time windows have the same effect on the DCA for this particular data set. The real-valued NSA with contant-sized detectors is not applicable to the data set. And the C4.5 decision tree algorithm provides a benchmark of the classification performance for this data set.

## 1 Introduction

Intrusion detection is the detection of any disallowed activities in a networked computer system. Anomaly detection is one of the most popular intrusion detection paradigms and this involves discriminating between normal and anomalous data, based on the knowledge of the normal data. Compared to traditional signature-based detection, anomaly detection has a distinct advantage over signature-based approaches as they are capable of detecting novel intrusions. However, such systems can be prone to the generation of false alarms. The Dendritic Cell Algorithm (DCA) is an Artificial Immune Systems (AIS) algorithm that is developed for the purpose of anomaly detection. Current research with this algorithm [6, 4] have suggested that the DCA shows not only excellent performance on detection rate, but also promise in assisting in reducing the number of false positive errors shown with similar systems.

To date, the data used for testing the DCA have been generated by the authors of the algorithm. While this approach provided the flexibility to explore the functionality of the algorithm, it has left the authors open to the criticism that the performance of the DCA has not been assessed when applied to a more standard data set. In addition to examining the performance of the DCA, such application allows for comparison with more established techniques.

For this purpose, the KDD Cup 1999 (KDD 99) data set [7] is chosen as the benchmark for evaluation, as it is one of the most widely used and understood intrusion detection data sets. This data set was originally used in the International Knowledge Discovery and Data Mining Tools Competition. During the competition, competitors applied various machine learning algorithms, such as decision tree algorithms [12], neural network algorithms [10] and clustering and support vector machine approaches [2]. In addition to these traditional machine learning algorithms, a range of AIS algorithms have been applied to this data set, such as real-valued Negative Selection Algorithm (NSA) [3].

The aim of this paper is to assess two hypotheses: Hypothesis 1, the DCA can be successfully applied to the KDD 99 data set; Hypothesis 2, changing the 'antigen multiplier' and the size of 'moving time windows' have the same effect on the DCA. We also include a preliminary comparison between the DCA, the real-valued NSA using constant-sized detectors (C-detector) and the C4.5 decision tree algorithm to provide a basic benchmark. This paper is organized as follows: Section 2 provides the description of the algorithm and its implementation; the data set and its normalization are described in Section 3; the experimental setup is given in Section 4; the result analysis is reported in Section 5; and finally the conclusions are drawn in Section 6.

## 2 The Dendritic Cell Algorithm

### 2.1 The Algorithm

The DCA is based on the function of dendritic cells (DCs) of the human immune system, using the interdisciplinary approach described by Aickelin et al. [1], with information on biological DCs described by Greensmith et al. [5]. The DCA has the ability to combine multiple signals to assess current context of the environment, as well as asynchronously sample another data stream (antigen). The correlation between context and antigen is used as the basis of anomaly detection in this algorithm. Numerous signal sources are involved as the input signals of the system, generally pre-categorized as 'PAMP', 'danger' and 'safe'. The semantics of these signals are shown as following:

- PAMP: indicates the presence of definite anomaly.
- Danger Signal (DS): may or may not indicate the presence of anomaly, but the probability of being anomalous is increasing as the value increases.
- Safe Signal (SS): indicates the presence of absolute normal.

The DCA processes the input signals associated with the pre-defined weights to produce three output signals. The three output signals are costimulation signal (Csm), semi-mature signal (Semi) and mature signal (Mat). The pre-defined weights used in this paper are those suggested in [5], as shown in Table 1. The equation for the calculation of output signals is displayed in Equation 1,

$$O_j = \sum_{i=0}^{2} (W_{ij} \times S_i) \ \forall j \quad (1)$$

here $O_j$ are the output signals, $S_i$ is the input signals and $W_{ij}$ is the transforming weight from $S_i$ to $O_j$.

|  | PAMP $S_0$ | Danger Signal $S_1$ | Safe Signal $S_2$ |
|---|---|---|---|
| Csm $O_0$ | 2 | 1 | 3 |
| Semi $O_1$ | 0 | 0 | 3 |
| Mat $O_2$ | 2 | 1 | -3 |

Table 1. Suggested weights for Equation 1

The DCA introduces individually assigned migration thresholds to determine the lifespan of a DC. This may make the algorithm sufficiently robust and flexible to detect the antigens found during certain time periods. For example, in real-time intrusion detection there are always certain intervals between the time when attacks are launched and the time when the system behaves abnormally. The use of variable migration thresholds generates DCs whom sample different time windows, which may cover the intrusion intervals.

An individual DC sums the output signals over time, resulting in cumulative Csm, cumulative Semi and cumulative Mat. This process keeps going until the cell reaches the completion of its lifespan, that is, the cumulative Csm exceeds the migration threshold, the DC ceases to sample signals and antigens. At this point, the other two cumulative signals are assessed. If the cumulative Semi is greater than the cumulative Mat value, the cell differentiates towards semi-mature state and is assigned a 'context value' of 0, and vice versa - greater cumulative Mat results in the differentiation towards mature state and a context value of 1. To assess the potential anomalous nature of an antigen, a coefficient is derived from the aggregate values across the population, termed the 'MCAV' of that antigen. This is the proportion of mature context presentations (context value of 1) of that particular antigen, relative to the total amount of antigens presented. This results in a value between 0 and 1 to which a threshold of anomaly, termed 'MCAV threshold', may be applied. The chosen value for this threshold reflects the distribution of normal and anomalous items presented within the original data set. Once this value has been applied, antigens with a MCAV which exceeds this threshold are classified as anomalous and vice versa. To clarify the algorithm a pictorial representation is present in Figure 1.

2.2 The Implementation

The general function of the system is to read data instances of the data set and then output the MCAV of each type of antigens. In order to implement this function, three major components are implemented:

– Tissue: processes the data source to generate antigens and signals, in each iteration Tissue stores the antigens into random indexes of an antigen vector and updates current signals to a signal vector.

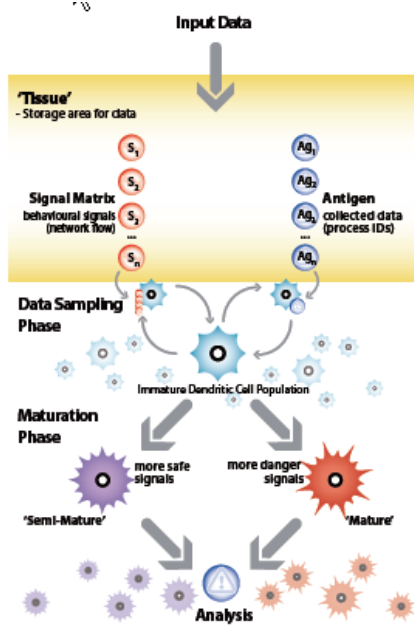

Fig. 1. The illustration of the DCA processes

- DCell: manages the DC population and interacts with Tissue to process the antigens and signals.
- TCell: interacts with DCell to produce the final results.

Two additional functions, antigen multiplier and moving time windows, are added into the system for the purpose of optimization. The DCA requires multiple instances of identical antigens, termed the 'antigen type', so processing across a population can be performed in order to generate the MCAV for each antigen type. The antigen multiplier is implemented to overcome the problem of 'antigen deficiency', that is, insufficient antigens are supplied to the DC population. As one antigen can be generated from each data instance within a data set such as KDD 99, the antigen multiplier can make several copies of each individual antigen which can be fed to multiple DCs.

The inspiration of applying moving time windows is from processes seen in the human immune system. The signals in the immune system persist over time, thus they can influence the environment for a period of time. The persistence of the signals can be presented by the cascade of signals within their affective time period. Due to missing time stamps in the KDD 99 data set, tailored window sizes for each data instance are not applicable, and a fixed window size is applied. The new signals of each iteration are calculated through Equation 2,

$$NS_{ij} = \frac{1}{w} \sum_{n=i}^{i+w} OS_{nj} \quad \forall j \tag{2}$$

```
input : antigens and pre-categorized signals
output: antigen types plus MCAV
initialize DC population;
while incoming data available do
    update tissue antigen vector and signal vector;
    randomly select DCs from DC population;
    for each selected DC do
        assign a migration threshold;
        while cumulative Csm<=migration threshold do
            get and store antigens;
            get signals;
            calculate interim output signals;
            update cumulative output signals;
        end
        if cumulative Semi<=cumulative Mat then
            cell context=1;
        else
            cell context=0;
        end
        log antigens plus cell context;
        terminate this DC and add a naive DC to the population
    end
end
while TCell analysis is not completed do
    for each antigen type do
        calculate MCAV;
    end
    log antigen types with corresponding MCAV;
end
```
Algorithm 1: Pseudocode of the implemented DCA.

where $NS_{ij}$ is the new signal value of instance $i$ in category $j$, **w** is the window size, and $OS_{nj}$ is the original signal value of instance $n$ in category $j$.

In brief the DCA combines multiple sources of input data in the form of pre-categorized signals and antigens. This input is processed across a population of DCs to produce the MCAV which is used to assess if an antigen type is normal or anomalous. Antigen multiplier and moving time windows are added to the algorithm to adapt the KDD 99 data set for use with this algorithm, as well as to assess the hypothesis of they having the same effect on the DCA. The pseudocode of the implemented DCA is shown in Algorithm 1.

## 3 The KDD 99 Data Set and Normalization Processes

### 3.1 The Data Set

The KDD 99 data set is derived from the DAPRA 98 Lincoln Lab data set [8] for the purpose of applying data mining techniques to the area of intrusion detection.

The DAPRA 98 data set contains two data sources, which are the network sniffer data from the sniffer placed between a router and the outside gateway and the Solaris system audit data from the Solaris audit host. The KDD 99 summarizes the two data sources into connections (data instances), each connection has 41 features (attributes), which can be grouped into four categories [11]:

- Basic Features: derived from the packet headers without inspecting the payload.
- Content Features: from the assessment of TCP packets by using domain knowledge of intrusion detection.
- Time-based Traffic Features: from the statistical analysis to captures the properties with a time window of two seconds.
- Host-based Traffic Feature: from the statistical analysis of the properties over the past 100 connections.

The KDD 99 is one of the few labeled data sets available in the field of intrusion detection. The data instances are labeled as normal connections or attack types, and the attacks can be grouped into four categories: Denial of Service (DOS), Remote to Local (R2L), User to Root (U2R) and Probe. The data set used in this paper is the 10% subset of the KDD 99 data set that is commonly used by other researchers. It consists of 494021 data instances, which are relatively massive. The whole data set would be more computational extensive, and hence much more difficult to handle, especially for the real-valued NSA with C-detector and the C4.5 decision tree algorithm. Both algorithms require training stage, the large the data set is, the longer the training would take. The 10% subset is statistically compared with the whole data set, and it features the similar ratio of the normal connections and the attacks.

3.2 Normalization of the Data Set

As anomaly detection is a two-class classification, the labels of each data instance in the original data set are replaced by either 'normal' for normal connections or 'anomalous' for attacks. Due to the abundance of the attributes, it is necessary to reduce the dimensionality of the data set, to discard the irrelevant attributes. Therefore, information gains of each attribute are calculated and the attributes with low information gains are removed from the data set. The information gain of an attribute indicates the statistical relevance of this attribute regarding the classification [11]. The information gain, termed Gain(S, A) of an attribute A relative to a collection of examples S, is defined as Equation 3 [13],

$$Gain(S, A) \equiv Entropy(S) - \sum_{v \in Values(A)} (\frac{|S_v|}{|S|} Entropy(S_v)) \qquad (3)$$

where *Values*(A) is the set of all possible values for attribute $A$, and $S_v$ is the subset of $S$ for which attribute $A$ has value $v$. The entropy of $S$ relative the

2-wise classification, termed *Entropy(S)*, is defined as Equation 4 [13],

$$Entropy(S) \equiv \sum_{i=1}^{2} -p_i log_2 p_i \qquad (4)$$

where $p_i$ is the proportion of *S* belonging to class *i*.

The histograms of the remainder attributes are assessed for the normalization of the DCA, to abstract the knowledge of both normal and anomalous. Based on the characteristics of the input signals, ten numeric attributes are grouped into the categories as follows:

- PAMP: serror rate, srv serror rate, same srv rate, dst host serror and dst host rerror rate.
- DS: count and srv count.
- SS: logged in, srv different host rate and dst host count.

Let x be the value of an attribute, if it is certain that anomalies appear when x [m, n], this attribute can either be PAMP or DS; otherwise if normality arises in this range, of this attribute is then normalized into the normalization defined by Equation 5,

$$f(x) = \begin{cases} 0 & x \in [0, m) \\ \frac{x}{n-m} \times 100 & x \in [m, n] \\ 100 & x \in (n, +\infty) \end{cases} \qquad (5)$$

where f(x) is the normalization function. The average of the multiple attribute values in each signal category is the value of that category. In addition, the other data steam of the DCA, the antigens, are created by combining three nominal attributes, which are protocol, service and flag. Multiple instances of each antigen type can generated through this way, which satisfies the requirement of the DCA for multiple observations of each antigen type. It makes sense in the case of both human immune system and intrusion detection: since antigens with the same pathogenic patterns can invade the human immune system over and over again; and attacks with the same patterns can be launched discretely over time in a networked computer system.

The ten attributes selected for the signals in the DCA are chosen to represent the detectors and antigens in the NSA. These attributes are normalized into the range from 0 to 1, using max-min normalization, thus the data space is a unitary hypercube $[0, 1]^{10}$. The data set is then rearranged to generate ten subsets through 10-folder cross-validation. The training data is made of the nine folders and the testing data is made of the one folder in each subset. The self set of the NSA is derived from all the normal data instances in the training data, and the antigens are the data instances in the testing data. The input data of the C4.5 decision tree algorithm contains the same attributes as those of the NSA but without normalization, and the labels of normal and anomalous are provided for the purpose of training.

## 4 Experimental Setup

Both the DCA and the NSA are implemented in C++ with the g++ 4.2 complier, and the C4.5 decision tree algorithm is performed in Weka [14], which is a collection of machine learning algorithms for data mining tasks. The experiments are run on a PC on which Ubuntu Linux 7.10 with a kernel version of 2.6.22-14-generic is installed. The receiver operating characteristics (ROC) analysis is performed to evaluate the classification performance of the DCA and the NSA. The true positive (TP) rate, false positive (FP) rate, true negative (TN) rate and false negative (FN) rate of each experiment are calculated, and the relevant ROC graphs are plotted as well. Three sets of experiments are performed: various DCA versions (E1), the real-valued NSA using C-detector (E2), and the C4.5 decision tree algorithm (E3).

In all experiments related to the DCA, the size of the DC population is set as 100 and it is constant as the system runs. The migration threshold of an individual DC is a random value between 100 and 300, to ensure this DC to survive over multiple iterations. The 'perfect MCAV' of an antigen type is calculated based on the labels of the original data set, normal is equivalent to context value 0 and anomalous is equivalent to context value 1. To generate the classification results of the DCA and the 'perfect classification results' from the perfect MCAVs, a MCAV threshold of 0.8 is applied. The MCAV threshold is derived from the proportion of anomalous data instances of the whole data set, which is equal to 80%. The classification results of the DCA are then compared with the perfect classification results, to assess the TP, FP, TN and FN. Three experiments of E1 are performed corresponding to the DCA versions as following:

- E1.1: the basic version of the DCA.
- E1.2: the system with antigen multiplier, the antigens are multiplied by 5, 10, 50 and 100.
- E1.3: the system with moving time windows, the window size is respectively equal to 2, 3, 5, 7, 10, 100 and 1000.

For each single experiment, ten runs are performed and the final result is the average of the ten runs. In order to make the results from different experiments more comparable, a fixed sequence of random seeds for ten runs is used. For E1.2 and E1.3, the two-sided Mann-Whitney test is performed to assess if various parameters can make the results statistically different from each other. The statistical significance a is set as 0.05, thus giving a confidence of 95% to either accept or reject the null hypothesis.

E2 includes a range of experiments of the NSA, as the data space increases from two dimensional to ten dimensional. According to the parameters mentioned in [9], the self radius is equal to 0.1 and the detector amount is increased to 1000 because of the large size of the data, and the matching rule used is the Euclidean distance matching. The results produced by the algorithm are compared to the labeled testing data, namely the 'perfect result', to perform the ROC analysis. The final results of each dimension is the average of ten subsets.

The experiment setup of C4.5 decision tree algorithm are as follows: the classifier chosen in Weka is J48, which is a class for generating an unpruned or a pruned C4.5 decision tree; the test option of the classification is set as 10-folder cross-validation.

| Category | Parameter | TP Rate | TN Rate | FP Rate | FN Rate |
|---|---|---|---|---|---|
| E1.1 | - | 0.7375 | 1 | 0 | 0.2625 |
| E1.2 | 5 | 0.75 | 1 | 0 | 0.25 |
| E1.2 | 10 | 0.74375 | 1 | 0 | 0.25625 |
| E1.2 | 50 | 0.75 | 1 | 0 | 0.25 |
| E1.2 | 100 | 0.75 | 1 | 0 | 0.25 |
| E1.3 | 2 | 0.75 | 1 | 0 | 0.25 |
| E1.3 | 3 | 0.75 | 1 | 0 | 0.25 |
| E1.3 | 5 | 0.74375 | 1 | 0 | 0.25625 |
| E1.3 | 7 | 0.75 | 1 | 0 | 0.25 |
| E1.3 | 10 | 0.75625 | 1 | 0 | 0.24375 |
| E1.3 | 100 | 0.71875 | 0.96 | 0.04 | 0.28125 |
| E1.3 | 1000 | 0.7 | 0.979592 | 0.0204082 | 0.3 |

Table 2. The ROC results of the experiments in E1

| Data Dimension | TP Rate | TN Rate | FP Rate | FN Rate |
|---|---|---|---|---|
| 2 | 0.98367 | 0.42944 | 0.37055 | 0.01633 |
| 3 | 0.23462 | 0.71834 | 0.08165 | 0.76538 |
| 4 | 0.08971 | 0.79289 | 0.00711 | 0.91029 |
| 5 | 0 | 0.79993 | 0.00007 | 1 |
| 6 | 0 | 1 | 0 | 1 |
| 7 | 0 | 1 | 0 | 1 |
| 8 | 0 | 1 | 0 | 1 |
| 9 | 0 | 1 | 0 | 1 |
| 10 | 0 | 1 | 0 | 1 |

Table 3. The ROC results of the experiments in E2

5 Result Analysis

The results of E1 are shown in Table 2, which indicate the antigen multiplier cannot consequentially enhance the system performance. The signals associated with the misclassified antigens are generated incorrectly from the original data set, thus the DCs always assign wrong context values no matter whether the antigens are multiplied or not. Moreover, the moving time windows cannot significantly improve the system performance either. Due to the limitation of the data set, the tailored window sizes of each data instance that may result in better system performance are not applicable. Furthermore, the Mann-Whitney test suggests a 95% confidence to accept the null hypothesis, that is, the results of all the experiments in E1 are not statistical different from each other.

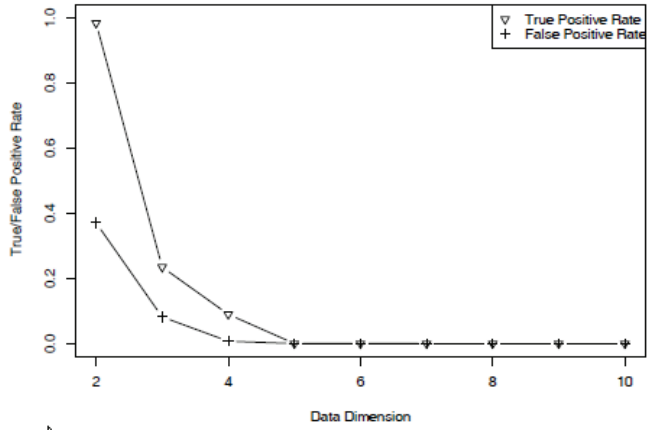

Fig. 2. Results of the real-valued NSA with C-detector across different dimensionality

The results of E2 are shown in Table 3, and the ROC results E2 from two dimensional to ten dimensional are shown in Figure 2. The algorithm produces acceptable results when the data space is two dimensional. But as the dimensionality increases, the classification performance is getting worse and worse. The algorithm cannot detect any anomalies when the data space is six dimensional or more. As the dimensionality of the data space increases, the search space grows exponentially, thus it is becoming more and more difficult to generate sufficient detectors that can effectively cover the space of non-self.

The ROC graph of the results in E1 and E2 when the dimensionality is ten is shown in Figure 3. The results of the DCA are located on the top-left corner of the graph, showing that all versions the DCA can successfully detect around 75% true anomalies over all actual anomalies as well as produce no or few false alarms. The real-valued NSA with C-detector cannot produce any useful results, as it fails to detect any anomalies. Moreover, as expected the C4.5 decision tree algorithm produces superb results, the true positive rate is 0.988 and the false positive rate is 0.008. This algorithm is designed specifically for the purpose of data mining, its classification performance is supposed to be better than the other two algorithms that are designed for the purpose of anomaly detection. But in terms of false positive rate, the classification performances of the DCA and the C4.5 decision tree algorithm are comparable with each other.

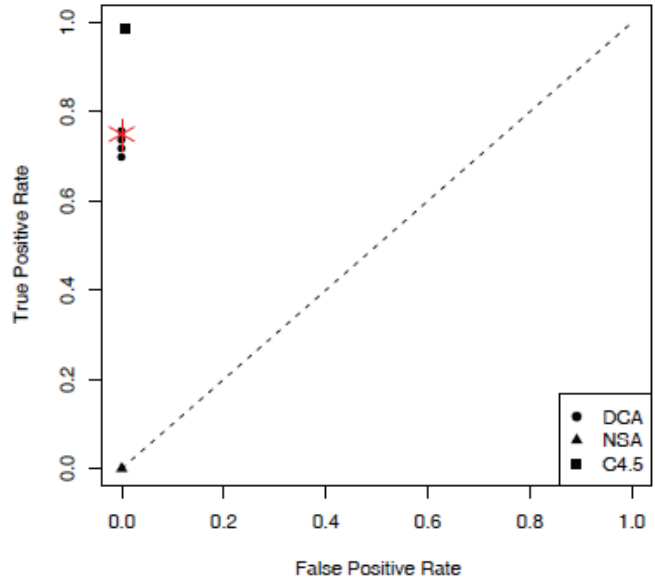

Fig. 3. The ROC graph of E1, E2 and E3 as data space is ten dimensional

## 6 Conclusions and Future Work

This paper presents the algorithm behaviors of the DCA, when it is applied to a standard data set, the KDD 99 data set. The results show that the DCA is able to work with the data set and produce reasonable performance, therefore Hypothesis 1 is accepted. Moreover, the DCA is an unsupervised learning algorithm, it does not require training with normal data instances. It acquires the knowledge of normal and anomalous through the categorization of signals based on basic statistical analysis. Besides, it is not constrained by high dimensionality of the data sets. Thus the DCA is applicable to large data sets with high dimensionality. The real-valued NSA with C-detector has poor classification performance on the high dimensional KDD 99 data set, it could not manage to detect any anomalies as the dimensionality increases up to six or more. Therefore, this algorithm is not applicable to the data sets with hight dimensionality. As a specialized machine learning algorithm, the C4.5 decision tree algorithm produces excellent results, it provides a benchmark showing the ideal results of the KDD 99 data set.

Due to limitations of the data set, the DCA could not be optimized by either antigen multiplier or moving time windows. First of all, it is only possible to generate one unique antigen from each data instance, leading to the insufficient observations of each antigen type by relative DCs, the problem cannot be solved with the antigen multiplier. Furthermore, the time stamps of each connection are unavailable, thus it is impossible to apply tailored window sizes in the system, and hence the advantage of the moving time windows is not fully utilized. Even though, both antigen multiplier and moving time windows have the same effect on the DCA for this particular data set, and hence Hypothesis 2 is accepted.

Some future directions of DCA research can be: first of all, to perform more rigorous comparisons between the DCA and other AIS algorithms; Secondly, to apply the DCA to other data sets, to further explore the limits of the DCA and to understand the antigen multiplier and moving time windows; Thirdly, to add more features to the DCA, to make the algorithm more adaptive and flexible.